\documentclass[12pt]{article}
\usepackage{amsmath}
\usepackage{times}
\usepackage{graphicx}
\usepackage{color}
\usepackage{multirow}
\usepackage[authoryear]{natbib}
\bibpunct{(}{)}{;}{a}{}{,}  
\usepackage{rotating}
\usepackage{bbm}
\usepackage{latexsym}

\newcommand{\captionfonts}{\normalsize}

\makeatletter  
\long\def\@makecaption#1#2{%
  \vskip\abovecaptionskip
  \sbox\@tempboxa{{\captionfonts #1: #2}}%
  \ifdim \wd\@tempboxa >\hsize
    {\captionfonts #1: #2\par}
  \else
    \hbox to\hsize{\hfil\box\@tempboxa\hfil}%
  \fi
  \vskip\belowcaptionskip}
\makeatother   

\begin{document}
\hspace{13.9cm}1


\noindent Final version of manuscript.\\  Accepted on January 24, 2026 for
publication in {\it Neural Computation.}

\vspace{19mm}

{\LARGE

  \noindent Autonomous Learning with \\
  High-Dimensional Computing Architecture \\
  Similar to von Neumann's}

\ \\
{\bf \large Pentti Kanerva} \\
Redwood Center for Theoretical Neuroscience \\
University of California at Berkeley, Berkeley, CA 94720-3198, USA \\
pkanerva@berkeley.edu
%


\thispagestyle{empty}
\markboth{}{NC instructions}
\ \vspace{-0mm}\\
%
\begin{center} {\bf Abstract} \end{center}

\noindent We model human and animal learning by computing with
high-dimensional vectors ($D$~= 10,000 for example). The architecture
resembles traditional (von Neumann) computing with numbers, but the
instructions refer to {\it vectors} and operate on them in {\it
superposition.} The architecture includes a high-capacity memory for
vectors, counterpart of the random-access memory (RAM) for
numbers. The model's ability to learn from data reminds us of deep
learning, but with an architecture closer to biology. The architecture
agrees with an idea from psychology that human memory and learning
involve a short-term working memory and a long-term data store.
Neuroscience provides us with a model of the long-term memory, namely,
the cortex of the cerebellum.  With roots in psychology, biology, and
traditional computing, a theory of computing with vectors can help us
understand how brains compute.  Application to learning by robots
seems inevitable, but there is likely to be more, including language.
Ultimately we want to compute with no more material and energy than
used by brains. To that end, we need a mathematical theory that agrees
with psychology and biology, and is suitable for nano\-technology. We
also need to exercise the theory in large-scale experiments. The
analogy with traditional computing suggests that the architecture be
{\it programmable} in terms of variables, values and data structures,
the very things that have made traditional computing ubiquitous and
that seem worth learning from and emulating.


\section {Preamble}

It seems best to clear a likely misconception up front. We generally
agree that the everyday computer is a ``von Neumann machine,'' and we
call other architectures ``non-von-Neumann,'' including Artificial
Neural Nets (ANNs). Computing with wide vectors, as discussed in this
paper, grew out of artificial neural nets and is an ANN of a kind, but
it realizes a von Neumann-like architecture with an instruction set
and a memory for wide vectors, and it allows variables and values to
be used in traditional ways. Three vector operations are used to
encode and decode composite structure, and to capture probabilities in
bit counts and vector similarity. Minimal reliance on backpropagation
is another thing that sets the ``new'' architecture apart from
mainstream ANNs. In abstract theory terms, the architecture is no less
powerful than an ordinary computer, or the Universal Turing Machine
without an endless tape. The finite automaton at the core is simple
and easily realized in an algebra of vectors. The point is to program
with vectors as basic objects, much as we now program with numbers.

\section {Background}

We explore the idea that human beings, animals, and robots can learn
to act intelligently in the world by building in memory a {\it
predictive model of the world.} The paper complements a 1988 report on
the Organization of an Autonomous Learning System, henceforth referred
to as the ``Target Article''
\citep{Kane-1988a}.

Neural-net associative memories for high-dimensional (aka ``wide'')
vectors were studied already in the 1970s, but operations of the right
kind for computing with vectors were found only in the 1990s, starting
with Plate's {\it Holographic Reduced Representation}
\citep{Plat-1991}.
Computing with vectors is subtle in that it takes place in {\it
superposition.}

We compare computing with vectors to conventional computing with
numbers and argue for modeling and simulating human and animal
cognition with a von Neumann-like architecture for computing with
vectors. We begin with an overview of conventional computing and
identify corresponding circuits for computing with vectors. We then
discuss reasons to distinguish between working memory and long-term
memory. The vector math is carried out in the working memory and the
results are stored in the long-term memory and ultimately drive
action. We make no attempt to map the working memory to neural
circuits in the brain, but observe that the circuits of the cerebellum
are laid out beautifully in 3D for a massive long-term memory for wide
vectors (see e.g.
\cite{Eccl+2-1967}
and figures 7.2 and 7.3 of
\cite{Llin+2-2004}):
the 3D layout resembles the wiring of magnetic-core memories of the
1960s, with their bit planes corresponding to the flat dendritic trees
of Purkinje cells.
\cite{Kane-1993}
illustrates step by step the cerebellum's likeness to computer memory
and to ANNs. The memory circuit consists of mossy fibers, granule
cells, Purkinje cells, and climbing fibers. The axons of granule cells
bifurcate into parallel fibers that run perpendicular to Purkinje
cells' flat dendritic trees.

The rest of the paper and the target article are about applying the
architecture to autonomous learning by robots, modeled after learning
by humans and animals with sensorimotor systems controlled by advanced
brains. The paper represents a systems approach that is aligned with
James Albus's ideas about robots with human-like faculties. His book
on {\it Brains, Behavior, and Robotics}
\citep{Albu-1981}
came out a decade before ideas about computing with vectors in
superposition began to take hold, but is still relevant and worth
reading.

\section {Traditional Computing with Numbers}

This section serves readers who know computers from having programmed
them in high-level languages but know less about the hardware within.
The lesson is that the computer in itself is exceedingly simple and
that its success comes from programming.

The traditional computing architecture, also known as the von Neumann
architecture, consists of a {\it memory} for storing programs and data
(Random-Access Memory, RAM), and a {\it processor} that steps through
the program and does the math (Central Processing Unit, CPU). The memory
is made of sequentially numbered {\it storage locations}, and a
location's number serves as its {\it address.} Even small computers have
memories with millions of storage locations.

A storage location is a piece of hardware that can hold a small amount
of data, for example a number, a letter, or a simple command for the
processor to carry out, such as adding two numbers. A storage location
can be visualized as a fixed number of bits, 32 for example, and the
programmer decides what the bits in any particular location---at any
particular address---mean.

The CPU has two functions, step-by-step {\it running} of the program and
the {\it math} that is built into the computer's circuits. The math
usually includes addition, subtraction, multiplication, and division of
numbers, and logic operations on strings of bits (AND, OR, NOT, shift
and rotate). More complicated operations are made with programs that
combine these basic operations. For example, a program to calculate the
mean of a set of numbers copies the numbers, one at a time, from the
memory into the processor, adds it to the sum, divides the final sum by
the count, and stores the result in memory for later use---for comparing
the mean heights of two groups of people, for example.

\subsection{Random-Access Memory---RAM}

{\it Random Access} is an all-important property of the memory. It means
that the contents of any storage location are immediately available to
the processor. The processor merely needs to ``know'' the location's
address. Random access is in contrast to storing data on tape or disk
that make the processor wait for the addressed memory location to be
under the read/write head. The ``miracle'' of random access is achieved
with an address-decoder circuit.

\subsection{Linked Lists}
\label{subsec:linked}

Random access allows lists of numbers to be stored so that any number
in the list can start the retrieval of the {\it rest} of the list. The
numbers in the list are used as addresses of storage locations and as
data to store, and the list is stored by stepping through it one
number at a time, and using it as the address in which to store the
{\it next} number. Linked lists (also called {\it pointer chains}) are
standard fare in list-processing languages such as Lisp, and provide a
way to retrieve a list starting with {\it any} number in it. When
vectors are linked in a list like that, they can become a predictive
model of the world, as discussed under ``predicting'' in
subsection \ref{subsec:sensing}.

Finally, computers get their inputs from the world and deliver their
outputs through transducers that convert between physical signals and
their representation in bits.

\section {Computing with Vectors}

By a $D$-{\it dimensional vector} we mean a series of $D$ numbers
treated as a coherent unit. The traditional (von Neumann) architecture
for computing with numbers is readily extended to computing with wide
vectors (e.g., $D$ = 10,000). The ``new'' architecture is essentially
the old: central processing unit (CPU) to carry out the vector math,
Random-Access Memory (RAM) for wide vectors, and circuits for input
and output. A traditional computer would control the running of the
program, with a vector-arithmetic unit running as a coprocessor, for
the time being at least.

\subsection {Experimenting with Vectors}

We wrote a program for identifying languages using four operations on
vectors: addition, multiplication, permutation, and dot
product
\citep{Josh+2-2017}.
The experiment was motivated by people's ability to identify languages
by how they sound, without knowing any of the words. Rather than
speech, we used about a million bytes of text for each of 21 European
Union languages, transcribed in Latin alphabet of 26 letters and a
symbol for the space between words. The challenge was to identify
languages by how they look in print, without knowing any of the words.
We made a 10,000-dimensional (10K) profile vector for each language
and for each test sentence, and identified a test sentence with the
language that had the most similar profile.

An analysis of the language profiles clustered them in four families:
Slavic, Baltic, Romance and Germanic, with English between Romance and
Germanic. Of 21,000 test sentences, 97.3\% were identified correctly,
and mistaken ones fell mostly in the same language family. The
experiment was ran on a laptop computer in a {\it single pass} over
the data and the test sentences, and took {\it less than 8 minutes}.

Profiles for languages and test sentences were computed with the same
algorithm, and were compared to each other with the cosine. We started
with 10K vectors for {\it letters,} computed 10K vectors for {\it
trigrams,} and added them into 10K vectors for {\it profiles}. Each of
the 27 letters was assigned a 10K independent random {\it seed vector}
of 1s and $-1$s. The trigrams of the data are all its three-letter
sequences and they were encoded into 10K trigram vectors with {\it
permutation} and {\it multiplication}: permute the first letter (its
10K code) twice, permute the second letter once, take the third as is,
and multiply the three. Permutation was done by rotating vector
coordinates and multiplication was done coordinate by coordinate.
Finally, a 10K profile vector was computed by {\it summing} over all
the trigram vectors of the text: the profile and test vectors were
{\it sums of products of permutations} of seed vectors for
letters. The meanings of ``add,'' ``multiply,'' ``permute,'' and
``similar'' are explained next.

\subsection {Instruction Set for Vectors}
\label{subsec:InstSet} 

What operations would make for useful computing with vectors? Linear
algebra perhaps, a well-established branch of mathematics that has a
major role in artificial neural nets. However, we want operations that
do not change dimensionality, so that an output of an operation can
serve as an input to further operations, and also as an address to
memory. However, the product of two vectors in linear algebra is
either a matrix or a number, and so we need a different way to
multiply vectors. We do it coordinatewise, known as the Hadamard
product.

Three simple operations make a surprisingly powerful system to compute
with: vector {\it addition,} coordinatewise {\it multiplication,} and
the shuffling of vector coordinates ({\it permutation}). The vectors
can be binary or integer or real or complex---the computing power
comes more from high dimensionality (e.g., $D$ = 10,000) and
randomness than from the nature of vector components.

The operations have been discussed at length in the literature
\citep{Plat-2003,Kane-2009,Kane-2023}
and can be illustrated with {\it binary} vectors (equivalent to
vectors of 1s and $-1$s). Addition is by coordinatewise majority rule
with ties broken at random (equivalently, ordinary vector addition
followed by applying a threshold to each coordinate); multiplication
is coordinatewise Exclusive-Or (XOR); and permutations reorder vector
coordinates. The algebra of addition and multiplication resembles
ordinary arithmetic in regard to associativity, commutativity,
invertibility, and distributivity. Permutations have no counterpart
in ordinary arithmetic. They are invertible, most do not commute, and
they distribute over both vector addition and multiplication.
Permutations of coordinates add all finite groups up to size $D$ into
the vector math.

The operations allow traditional data structures (sets, sequences,
lists, trees, stacks, etc.) to be represented and operated on in
superposed vectors
\citep{Kley+10-2022}.
For example, {\it sets} can be encoded with addition, the {\it
binding} of variables to values with multiplication, {\it sequences}
with permutations, and data structures at large with combinations of
the three operations. For a simple example, a vector that means $x =
a$ can be encoded by the product $P = X*A$ where $X$ and $A$ are
$D$-dimensional vectors representing the variable $x$ and the value
$a$, and * is coordinatewise multiplication (XOR for binary
vectors). The three operations are also used to decode a composed
vector into its constituent vectors: the value of $x$ in $P$ can be
recovered by multiplying $P$ with (the inverse of) $X$, namely $X*P =
X*(X*A) = (X*X)*A = A$, and by noting that XOR is its own inverse so
that $X*X$ evaluates to a vector of 0s. In ordinary computing, the
variable $x$ is represented by an address of a memory location and is
buried in the program code, and the value $a$ is represented by the
contents of the location. The three operations also allow regularities
in data to be captured in vectors, opening a path to statistical
learning.

A fourth operation, based on the dot product, measures the {\it
similarity} between vectors (Hamming similarity, cosine, or Pearson
correlation for example). The idea is that when vectors are similar
their meanings are similar. Nearly all pairs of vectors in a
high-dimensional space are dissimilar (effectively
orthogonal)---randomly drawn pairs of vectors are approximately
orthogonal. An endless supply of approximately orthogonal vectors is a
major reason for insisting on high dimensionality.

\subsection {Computing in Superposition}

Computing in superposition goes by many names, depending mainly on the
intended audience---whether mathematicians, physicist, cognitive
scientists, linguists, engineers, or computer scientists: Holographic
Reduced Representation (HRR), Vector Symbolic Architecture (VSA),
Context-Dependent Thinning, Hyperdimensional Computing, and Semantic
Pointer Architecture (SPA). Each item or element or feature or thing
or piece of data---each vector---that is included in a superposed
vector is distributed evenly over all vector components---any subset
of vector components represents the same thing as the entire vector,
only less precisely. Representation of this kind is called
``holographic'' or ``holistic,'' as opposed to the conventional
localist representation where a variable is represented by an address
of a memory location and its value by the contents of the location.
The encoding of $x = a$ with the vector $P$ in
subsection \ref{subsec:InstSet} is an example of holistic
representation: $P = X * A$.

Computing in superposition begins with $D$-dimensional {\it seed
vectors} chosen at random (with i.i.d. components), to represent basic
entities such as variables and names. In essence, they are {\it
symbols} or {\it tokens} to compute with. They can be used as
addresses to memory and as data stored in memory, and as inputs to the
math operations that combine them into new vectors, which in turn can
address the memory, be stored as data, and serve as inputs to the
vector operations. New representations are made from existing ones
with explicit calculation, which is fundamentally different from the
generation of representations in an autoencoder or in the layers of a
deep neural net as it is trained. Distributivity and invertibility of
operations also mean that codes for composite objects can be decoded
into their constituents, in some cases at least.

The operations allow a high degree of {\it parallelism.} Because
addition and multiplication happen coordinatewise, independently of
other coordinates, all $D$ (= 10,000) can be done at the same time.
Furthermore, because the operations are simple, it is feasible for
each dimension to have its own processor circuit. Computing with
vectors in superposition can then be seen as each coordinate having
its own simple computer, all computing the same thing but at
exceedingly low precision---down to 1-bit. On the aggregate, however,
they compute with high precision if the dimensionality is high
enough---and even if some of the simple circuits malfunction! In
contrast, traditional circuits for computing with numbers are
complicated and are expected to work flawlessly. The simplicity of the
coordinatewise operations makes it possible to distribute them
throughout the system, reducing the need to move data before it can be
computed on. This is called ``in-memory computing.''

\section {Ideas for Architecture from Biology and Psychology}

How does computing with vectors fare in the light of biology and
behavioral sciences? Psychological research in the 1960s and '70s
identified two levels of organization of human memory, short-term
working memory and long-term data store
\citep{Atki+Shif-1968, Ande+Bowe-1973};
see also Wikipedia article on ``Atkinson--Shiffrin memory
model''). Representations are assembled and analyzed in the working
memory and stored in the long-term memory
\citep{Tulv+Dona-1972, Tulv+Thom-1973}.
These early models summarize behavior without accounting for
underlying mechanisms and their math.

The psychologists' models are similar to traditional computing that
assembles and interprets data structures in the central processing
unit (CPU) and stores them in the random-access memory (RAM).
Computing with vectors follows the same organization: vectors are
encoded and decoded in the working memory (counterpart of CPU) and
stored indefinitely in the long-term memory (counterpart of RAM). We
will discuss the long-term memory first.

\subsection {Long-Term Memory for Vectors}

Memory allows past experience to bear on the present situation. {\it
Associative memory,} also called {\it content-addressable} (these
terms lack a precise definition), has been an object of psychological
inquiry for over a century
\citep{Jame-1890, Raai+Shif-1981},
and mathematical models have been proposed since the early days of
artificial neural nets
\citep[see][for a review]{Hint+Ande-1981}.
Sparse Distributed Memory
\citep{Kane-1988b}
takes the idea a step further by assuming that the memory could have
billions of locations and that its function is long-term data
storage. In the rest of this paper, ``long-term memory,''
``associative memory,'' ``high-dimensional RAM,'' ``Sparse Distributed
Memory,'' and ``SDM'' mean the same thing and are often referred to
simply as {\it the memory}. Its similarity to computer memory is
significant: the computer RAM stores numbers and is addressed by
numbers, an associative memory such as the SDM stores vectors and is
addressed by (cued with) vectors.

Psychologically realistic models of memory for vectors, such as the
Sparse Distributed Memory
\citep{Kane-1988b, Kane-1993},
consists of locations for storing $D$-dimensional vectors. Each
location is associated with a $D$-dimensional address. A {\it location
is activated} when the {\it memory is probed/cued/addressed} with a
vector that is {\it similar} to the location's address. The similarity
criterion is chosen so that multiple locations are activated, but only
a tiny fraction of all possible locations
\citep{Kane-1993}.
An activated location either accepts a $D$-dimensional vector (write
operation) or outputs a $D$-dimensional vector (read operation). The
final output of a memory read is computed from the outputs of all
activated locations. Although the number of possible locations borders
the incalculable, it is possible to {\it simulate random access} by
distributing data over multiple locations and reconstructing it upon
retrieval.

There is a difference of lesser importance to note: storing a number
in a computer memory replaces the old contents of the
activated/addressed location; storing a vector in an associative
memory {\it adds it to the contents of all activated locations}---the
newly stored data are {\it superposed.}

An associative memory's likeness to the cortex of the cerebellum is
remarkable
\citep{Marr-1969, Albu-1971, Kane-1993},
and the numbers agree with the need for a high-capacity memory. With
50 billion granule cells
\citep{Llin+2-2004}
representing locations for storing of vectors, the human brain could
store one vector every minute (more than 10 million bits a day) over a
life span of a 100 years.

{\it Catastrophic forgetting} is a known phenomenon of artificial
neural nets. It is in part due to the limit on the amount of
information that any given ``hardware'' can store, and in part to
training a monolithic network with error back-propagation
\citep{Fren-1999}.
Catastrophic means that when the capacity limit is reached, the
net begins to produce outputs that are unrelated to the vectors it was
trained on---it's like a phase transition in physics where the
properties of a substance change suddenly, as when super-cooled water
turns into a block of ice. The alternative is {\it graceful
degradation}, which can be achieved with a memory that is organized in
storage locations whose contents can change but addresses are fixed,
as they are in ordinary computer memory, in the Sparse Distributed
Memory, and apparently also in the cortex of the cerebellum.
Parallel-fiber synapses with Purkinje-cell dendrites are modifiable
and make up over half the brain's synapses;
\citep{Llin+2-2004}.
However, a small fraction of synapses (1\%) should be fixed according
to the theory, namely, synapses between mossy fibers and granule cells
that determine which locations/granule cells to activate. They act as
an address decoder and provide a framework for recording a lifetime of
experience.

\subsection {Short-Term Working Memory}

There is more to consider when computing with vectors in superposition:
just how many vectors to encode into a single vector? The number is
limited by vector dimension and determines the amount of structure that
a single vector can hold
\citep{Gall+Okay-2013, Frad+2-2018},
beyond which it is necessary to store vectors in the long-term
memory. Psychologists call it {\it chunking.}

The human mind can recall large amounts of data that has been
organized in chunks (and in chunks of chunks, and so on) and stored in
memory. George
Miller's ``Magical Number Seven, Plus or Minus Two'' is the classic on
the topic
\citep{Mill-1956}.
The working memory combines pieces of data into chunks and stores them
in the associative memory, and recalls chunks into the working memory,
to compose them into further chunks or to break them down into their
constituent parts.

Miller's magical number suggests that a chunk should combine fewer
than ten items. That is also about the number of vectors that can be
encoded into a single vector and readily decoded with the vector math.
The number grows nearly linearly with vector dimensionality $D$
\citep{Frad+2-2018}.
With psychology and the vector math as clues, we assume that a
10,000-dimensional working memory can hold and operate on up to ten
vectors at a time.

\section {Modeling the World with Vectors}

To help us understand intelligence of animals (and of ourselves), we
liken them (and us) to robots whose behavior is controlled by
``brains'' that compute with wide vectors.
\cite{Albu-1981}
makes a case for humanlike robots, and
\cite{Kane-1988b}, \cite{Plat-1991} and \cite{Pian+6-2024}
argue for representing concepts with high-dimensional vectors. The
robot has sensors and actuators for interacting with the world, an
associative memory for storing a model of the world, and a working
memory where the world model is constructed and interpreted with the
vector math
\citep{Teet+3-2023}.

\subsection {The Focus}

The contents of the working memory are summarized in a single
$D$-dimensional vector called the {\it focus}. The idea of a focus at
the core of a cognitive computing architecture goes back to 1988
\citep{Kane-1988a, Kane-1988b}
but has not been adopted widely. However, it deserves a new look now
that von Neumann-like computing with vectors in superposition has
become understood
\cite{Plat-1991}.
The sensors feed into the focus, the activators are driven from the
focus, the memory is addressed by the focus, and vectors in and out of
the memory pass through the focus. The vector in the focus is the
system's (robot's) {\it state} and it represents the robot's
subjective experience at the moment---the focus is the robot's
subjective self (an objective state would have to include the contents
of memory as well).

The focus is a theoretical construct, meant to capture the idea of an
{\it integrated self}. For example, any of a number of things can bring
to mind a specific experience from the past; the trigger can be a
specific sight or sound or smell or ache or pain, or some combination of
them. Thus any one of them can serve as an address to memory, to evoke
the past experience. Similarly, we recognize a person by how they look
or sound or move or act: all those different modalities refer to---and
retrieve from memory---the same person.

It seems befitting that the focus is limited to a single vector
because our minds can attend to only a few things at a time
(cf. Miller's magical number 7). Limited capacity is behaviorally
realistic. Furthermore, because the focus is at the crossroads of
sensors, actuators, and memory, an experience created by the world
through the senses can also be created from the memory. In computing
terms, they reside in a {\it common $D$-dimensional space,} allowing
the history to be stored as a linked list; see
subsection \ref{subsec:linked}.

As an added benefit, representing sensory modalities and action in
common mathematical space allows generic algorithms to be used for
learning. It also means that we need to turn physical (and chemical)
signals with very different properties (sight, sound, taste, smell,
temperature, pressure, etc.) into $D$-dimensional vectors for the
working memory. The best ways to do it will be worked out over time.
Meanwhile, standard signal-processing techniques together with random
projections have been used successfully. For example, letters of the
alphabet have been represented by random vectors for written-language
identification
\citep{Josh+2-2017},
words of a language have been
represented by random vectors for comparing word meaning
\citep{Kane+2-2000},
and Fourier coefficients for speech have been randomly projected to a
$D$-dimensional space for speaker identification
\citep{Huan+4-2022}.

Looking ahead, we need to learn how nervous systems and brains prepare
sensory input for further processing, and build that into artificial
systems. For example, the cochlea of the inner ear analyses sound into
frequencies---it Fourier-transforms the sound before passing it on to
the rest of the brain. Regarding vision, the optic nerve brings in
information along about 1.2 million axons per hemisphere
\citep{Ster+Demb-2004}
and the primary visual cortex distributes it among some 140 million
neurons
\citep{Leub+Kraf-1994}---a
100-fold increase.

\subsection {The Passage of Time}

To act intelligently in the world, the actor---an individual or a
robot---must deal with the passage of time. We are concerned here with
subjective time and represent it with a sequence of {\it moments,} and
a moment with the focus vector. A system's {\it history} is then a
life-long sequence of vectors for moments of time. The history can be
divided into {\it episodes} defined by shared context. The challenge
is to store the history in a way that allows a system to learn from
experience.

Short sequences, up to ten or so, can be encoded into a single vector
in the working memory and treated as a chunk.
\cite{Josh+2-2017}
encoded short sequences of letters ($N$-grams) as chunks and used them
to identify languages. The long-term associative memory makes it
possible to deal with longer sequences by storing the history in the
long-term memory as a life-long linked list; see
subsection \ref{subsec:linked}. The vector for the present
moment---the present state, the focus---can then be used as an address
to trigger the recall of similar past episodes.

\subsection {Sensing and Acting}
\label{subsec:sensing}

{\it Sensing.} Animals have elaborate sensory systems that
receive data about the world: the brain ``swims'' in a torrent of data
and needs to find regularities and patterns in the data stream. A
wealth of sensory input comes from the body itself. In fact, the body
can be regarded as a part of the world from which we receive input and
on which we act.

{\it Detecting.} Recognizing previously encountered states makes it
possible to detect irregularities and anomalies that can serve as an
alarm, for example. It does not take much to see that ``wrord'' is
anomalous, for example.

Detection can be accomplished with very basic learning: store the
vector for each moment in the memory using the vector itself as the
address---this is called ``autoassociation.''  Frequently occurring
similar vectors are reinforced and infrequent vectors become
background noise. When the sensory data for a moment arrives, it can
be used to address the memory to determine whether it represents
something new or something ``seen'' before. If it recalls a dissimilar
vector the new vector is stored, if the two are essentially the same
the new vector can be ignored, but if they are similar, the difference
between the two can be analyzed with the vector math to choose an
action to take.

{\it Predicting.} Intelligent behavior requires more than the ability
to detect whether I have seen this before or been here before. We need
to look ahead: what to expect next and how to act accordingly. We need
to exploit regularities in data that unfold over time.

Prediction can be accomplished by storing the system's history in the
memory as a linked list, as described in
subsection \ref{subsec:linked}. When a vector for a moment recurs and
is used to address the memory, it recalls---and predicts---the next
moment. A moment later the prediction can be compared to what actually
happens---this is essentially anomaly detection that can initiate
further action.

{\it Acting.} To talk of acting implies the existence of an {\it
actor.} We liken it to a robot or an animal or a human being coping
with a seemingly boundless and indifferent world. To exist in the
world, the actor needs to acquire resources from the world (material
and energy) and to steer out of harm's way---to eat and not to be
eaten. The idea of an actor implies the notion of ``self,'' an
individual that is separate from the rest of the world---we use the
word ``individual'' for actors of different kind but not for the world.
Individuals/actors have some degree of autonomy. The individual's
actions become a part of the world model.

{\it Reacting.} Reaction is a response to a stimulus. Some animal
behaviors are passed on genetically and other are learned during life.
The model deals with both---see discussion of the {\it preference
function} in the target article. Learning in real time takes advantage
of the memory's ability to predict. The actions and reactions of
today's robots are mostly programmed, and very little learning takes
place in real time out in the field.

{\it Interacting} implies two or more parties reacting to each other's
actions. If one of the parties is the indifferent world, the
individual can act on it, to discover how the world works. When the
interaction is between two individuals, each one can learn about the
other by probing, but there is more: rather than being indifferent,
the two can cooperate or compete, in which case my model of the world
needs to include a model of you, which needs to include a model of me,
and so on. To avoid infinite regress, we project ourselves to the
other individual---act as if the other were like ourselves---and
regress only a step or two beyond that.

What is the individual's relation to its model of the world: how is
the model used?  We can think of the model as a database that the
individual consults, to evaluate alternative actions and to choose one
that is appropriate for the present situation. Most of today's robots
work like that, but it is not what actually happens in brains because
it assumes a high-level executive (a homunculus) that runs the show.
The use of the model by the individual needs to be more organic.
Looked at in that way, rather than an individual having a model of the
world, the individual {\it is} the model (and the model is the
individual).

Interaction with the world and its individuals are the data for the
world model. An individual's interaction is experienced as a sequence
of states of the focus that include both sensory and motor components.
When stored in memory as a linked list, it allows the individual to
predict and to act according to the prediction. The details are
worked out in the target article
\citep{Kane-1988a}
and will not be repeated here. We merely comment on it.

\section {Autonomous Learning (Kanerva 1988a) \\
in Retrospect}

Computing with wide vectors tries to capture the organization of animals
that learn and adapt. It is based on the idea that an associative memory
can learn {\it predictive} modeling of the world with a circuit that is
strikingly similar to the cerebellum's. In psychologists' terms, the
associative memory is a long-term data store.

The architecture includes a working memory. Its contents are
summarized in a single vector called the {\it focus.} The state of the
focus represents the system's/robot's/indi\-vidual's subjective
experience from moment to moment.

The 1988 paper has nothing concrete to suggest about how the working
memory actually works. The paper merely refers to an ``Encoding
Problem'' as yet to be solved. The gist of the encoding problem was
solved by Plate by computing in {\it superposition} with Holographic
Reduced Representation
\cite{Plat-1991}.
Other than that, the architecture is conventional (von Neumann) and
suitable for both symbolic
\citep{Gayl-2003}
and statistical computing
\citep{Josh+2-2017}.

The 1988 paper refers to varying the amount of information that a
given data source or destination contributes to the focus vector, but
does not tell us how actually to do it. Superposition suggests an
answer. When each data source and destination is encoded into a
$D$-dimensional vector, the vectors can be weighted by importance
before adding them to the focus. For example, we can prioritize sight
over sound, or vice versa, depending on the context. We would
no longer refer to feature vectors, as that suggests vectors being
partitioned into fields for features. Every ``feature'' that is encoded
into a vector is present in every coordinate. In fact, what we
traditionally think of as features are more like afterthoughts, to
help us compartmentalize the world. If everything in the world were
green, color as a feature would not exist.

\section {What Next?}

Computing with wide vectors seems particularly suited for
simulating---and possibly understanding---intelligence of animals with
advanced sensorimotor systems and brains
\citep{Albu-1981, Kane-1988b, Pian+6-2024}.
Much needs to be worked out, however, to develop the idea into a
viable technology. Fortunately, developing and testing does not
require excessive hardware or computing. Ample opportunity to
experiment, innovate and invent is within reach
\citep[see][for an example]{Yang-2023}.
Here are some of the challenges we face.

{\it Programming and Control.} Whether and how to do the actual
programming with superposed vectors is unclear. For now, conventional
programs running in conventional computers control the operation of
the hyperdimensional computer in simulation or proto\-type. However,
people are beginning to think also about vector-based control
structures
\citep[e.g.]{Tomk+Kell-2024}.

{\it Sparse Vectors.} Matching sets of operations for dense vectors
have been known from the start (1990s). They are easy to compute with
but inefficient. We need good sets of operations for sparse vectors
without compromising computing power. The algebra of the operations
must allow multiple vectors to be encoded into a single vector,
operated on, and the result decoded into known vectors.

{\it Associative Memory.} Computing with vectors depends crucially on
a high-capacity associative memory for wide vectors (a
high-dimensional RAM), that does a fast near-neighbor search in a huge
and sparsely populated address space. When vectors are stored in an
associative memory with themselves a addresses (called
autoassociative) they can become fixed points with basins of
attraction of a dynamical system, allowing rapid storing and
retrieving of vectors. The Sparse Distributed Memory (Kanerva 1988b)
demonstrates the concept but is inefficient. The circuits of the
cerebellum can suggest more efficient designs, for example by
judicious use of randomness and sparseness
\citep{Jaec-1989}.

{\it Noise.} One of nature's marvels is how we recognize a person or
an object or a place or a situation---anything---as being the {\it
same} now as at some earlier time, and distinct from other things, in
spite of the data arriving at the senses never repeating exactly. The
tendency to categorize things based on what we expect of them, seems
to be built in. But even ``built in'' needs explaining: a physical
process that sustains the behavior. Without going into it here, we
merely presume that high dimensionality plays an important role, and
instead comment on how variability in input is handled in cognitive
modeling today.

Objects, etc., are treated as points in a feature space that has a
similarity measure. Prototypical objects occupy regions of the space,
and nearby points are considered to be noisy versions of them. This
works when most points are far away, as they are in a high-dimensional
space. If the prototypes are stored in memory as fixed point, a noisy
version that falls in a basin of attraction is recognized with a
slight delay typical of attractor networks.

Another source of ``noise'' cannot be ignored. It has to do with
encoding and decoding of superposed vectors. For binary vectors,
multiplication and permutation are invertible and preserve
information, but addition is only approximately invertible, and so the
decoding of a vector sum puts out noisy vectors. That limits the
number of vectors included in a sum, and calls for cleaning up of
vectors decoded from a sum before further operating on them.

{\it Neuroscience.} We have a reasonable idea of how a long-term
memory in the brain is organized
\citep[cf. the  cerebellum:] {Albu-1971, Kane-1993, Llin+2-2004, Jaec-1989}
but none for the working memory. If computing with vectors produces
human-like or animal-like behavior with energy comparable to brain's,
we can hope to understand and identify brain's circuits for working
memory.

{\it Psychology and Psychophysics.} Psychologists have probed human (and
animal) behavior in countless experiments, and conjectured about
processes that underlie memory recall, response time, attention span,
interference, ability to discriminate, and so on. Relating the
observations to computing with wide vectors can deepen our understanding
of the underlying processes and lead to the engineering of more lifelike
robots, for example.

Modeling of brains with wide vectors also opens new vistas in
experimental psychology. As the theory and models improve, they will
suggest experiments to confirm or counter hypothesized underlying
mechanisms. This can benefit both psychology and engineering.

{\it Language} and {\it logic} are the apex of symbolic behavior and
have even been considered as defining intelligence. Symbolic
processing also formed the early core of AI---until we tried to build
systems that learn by interacting with the world. Where logic and
language failed, artificial neural nets were thought to succeed,
leading to today's deep neural nets and large language models.
Impressive as they are, their energy requirements so exceed the
efficiency of brains, as to rule them out as models of how brains work
and make us intelligent.

Human and animal intelligence is based on both learning from sensor
data---exploit\-ing regularities in the statistics---and reasoning,
even in the absence of fully developed language. Artificially
intelligent systems need to be capable of both, which is what
computing with vectors aims at. The algebra of the operations makes
symbolic processing possible (e.g., the explicit encoding, storage and
decoding of structured data), and high-dimensional representation can
include probabilities without explicit tallying and bookkeeping.

\section {Summary}

The paper outlines a traditional (von Neumann-like) architecture for
computing with wide vectors and argues for its use for autonomous
learning in real time by systems such as robots. It is also meant to help
us understand how humans and animals learn. It combines ideas from
psychology (short-term working memory and long-term data store,
chunking), neuroscience (cerebellum-like circuit for an associative
memory), computer science and engineering (central processing unit,
random-access memory, sequence as a linked list), and mathematics
(algebra of vector operations, effective orthogonality of random
vectors). Peculiar about computing with vectors is that it takes place in
superposition. Variables and values are vectors of a {\it common
mathematical space,} and so is a variable having a certain value. The
memory for storing the vectors is also addressed by the vectors---it's a
``high-dimensional RAM.''

The subjective state of the system is defined by a single vector
called the {\it focus.} It summarizes the contents of the working
memory. The sensors feed into the focus, the actuators are driven from
the focus, and the memory is accessed through the focus. The passage
of time is represented by a sequence of vectors in the focus. When
stored in the memory as a linked list, it becomes a predictive model
of the world.

Autonomous learning requires that some sensory states start out as
inherently desirable (satisfy a need) and some as inherently
undesirable (painful). A model of the world that includes the system's
own actions can then be used to learn sequences of actions that lead
to desirable states and avoid harmful ones.

The architecture of the long-term memory (Sparse Distributed Memory) is
{\it shallow.} It has {\it one fixed layer} for activating memory
locations and {\it one variable layer} for storing and recalling data in
the activated locations. The fixed layer corresponds to the
address-decoder circuit of a computer memory (RAM), and the variable
layer corresponds to memory stacks that store the data. This
architecture is suggested by the wiring of the cortex of the cerebellum,
which contains over half the brain's neurons and synapses. The variable
layer can be trained in real time by the perceptron learning rule, and
learning can happen in only a few rehearsals.

The vector math is carried out in the working memory (counterpart of
the Central Processing Unit, CPU) with an instruction set consisting
of coordinatewise addition and multiplication, and the permutation of
coordinates. Similarity of vectors is based on the dot product. The
operations can combine symbolic processing and statistical learning.

Interaction with the world happens through sensors and actuators.
Input is made into wide vectors for the working memory/focus with
sensor-specific {\it pre-processing,} and the focus is decoded for
commands to actuators with motor-specific {\it post-processing.} Pre-
and post-processing can be engineered, inherited or learned early in
the system's existence in multiple passes with backpropagation, and
then fixed.

Computing with vectors addresses many of the same issues as deep
neural nets but with an architecture more like that of a traditional
computer, or of animals with advanced brains. It should do well at
simulating human and animal cognition, but that is for future research
to determine.

\section*{Supplemental Material}

RIACS Report 88.14 on the Organization of an Autonomous Learning
System
\citep{Kane-1988a},
referred to here as the {\it Target Article}.


\subsection*{Acknowledgments}

This work was supported in part by CoCoSys, one of the seven centers
sponsored by the Semiconductor Research Corporation (SRC) and DARPA
under the Joint University Microelectronics Program 2.0 (JUMP 2.0);
and in part by the U.S. National Science Foundation (NSF) under
Cooperative Agreement No. 2433429, ``NSF Al Research Institute on
Interaction for Al Assistants (ARIA).''

Sincere thanks to professor Peter Denning for discussions that lead to
writing of the paper, and to an anonymous reviewer for comments that
eventually lead to its publication. Special thanks to professor Bruno
Olshausen and his Redwood Center at UC Berkeley, for creating an
environment favorable for scientific research and discovery.

\vspace {.5em} \noindent {\bf \large Publisher's Note.} All claims
expressed in this article are solely those of the author and do not
necessarily represent those of their affiliated organizations, or
those of the publisher, the editors and the reviewers. Any product
that may be evaluated in this article, or claim that may be made by
its manufacturer, is not guaranteed or endorsed by the publisher.


\bibliographystyle{APA}


\typeout{get arXiv to do 4 passes: Label(s) may have changed. Rerun}

\end{document}